\documentclass[runningheads]{llncs}

\usepackage{eccv}

\usepackage{eccvabbrv}

\newcommand{\figlabel}{Figure\xspace}

\newcommand{\tbllabel}{Table\xspace}
\newcommand{\innersection}[1]{\vspace{3pt}\noindent\textbf{#1}}

\newcommand{\supp}{\textbf{Appendix}\xspace}

\usepackage{todonotes}

\usepackage{listings}

\definecolor{Highlight}{HTML}{fc8d62}  

\usepackage{graphicx}
\usepackage{wrapfig}
\usepackage{booktabs}
\usepackage{multirow}
\usepackage{enumitem}

\usepackage[accsupp]{axessibility}  

\usepackage[pagebackref,breaklinks,colorlinks]{hyperref}

\usepackage{orcidlink}

\makeatletter
\def\thanks#1{\protected@xdef\@thanks{\@thanks
        \protect\footnotetext{#1}}}
\makeatother
\begin{document}

\title{FastPCI: Motion-Structure Guided Fast Point Cloud Frame Interpolation}

\titlerunning{FastPCI}

\author{Tianyu Zhang\inst{1}  \and
Guocheng Qian\inst{2}\orcidlink{0000-0002-2935-8570} \and
Jin Xie\inst{3\dag} \and Jian Yang\inst{1}
\thanks{$\dag$ Corresponding author.}}

\authorrunning{T.~Zhang et al.}

\institute{Nankai University, Tianjin, China \and
Snap Research \and 
State Key Laboratory for Novel Software Technology, Nanjing University, Nanjing, China
School of Intelligence Science and Technology, Nanjing University, Suzhou, China \\
\email{tianyu.zhang@mail.nankai.edu.cn}, \email{guocheng.qian@kaust.edu.sa} \\
\email{\{csjxie, csjyang\}@njust.edu.cn} }

\makeatletter
\let\@oldmaketitle\@maketitle
\renewcommand{\@maketitle}{\@oldmaketitle
\myfigure\bigskip}
\makeatother
\newcommand\myfigure{%
  \makebox[0pt]{\hspace{12.5cm}\includegraphics[trim=0cm 0cm 0cm 0cm, clip, width=1.0\columnwidth]{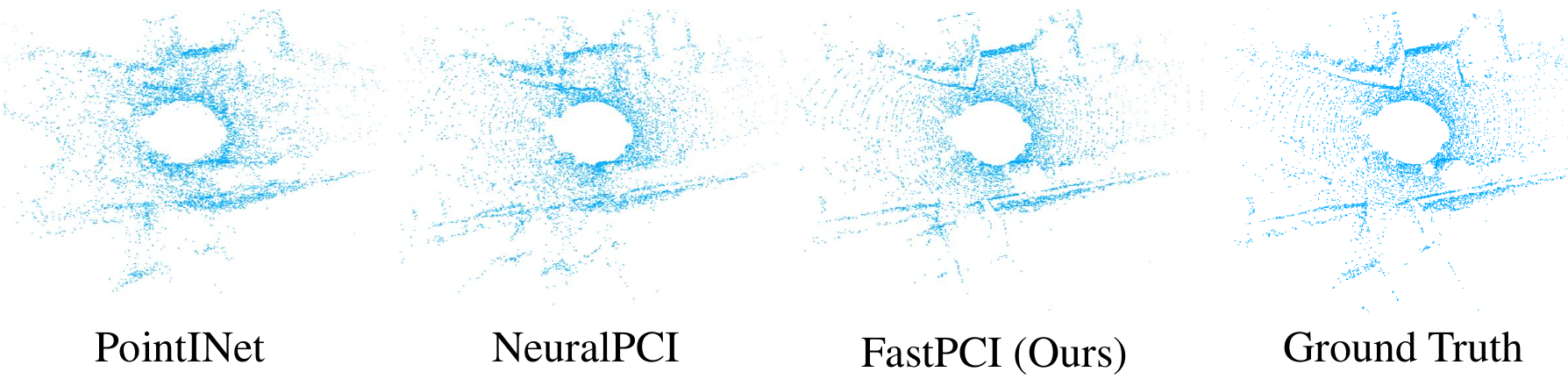}}
  \\
\refstepcounter{figure}\textbf{Fig. \thefigure}: Our FastPCI performs accurate point cloud frame interpolation in $\sim0.1$ seconds per frame, being more accurate and more than $600\times$ and $10 \times$ faster than the state-of-the-art NeuralPCI \cite{NeuralPCI} and PointINet \cite{PointINet}.
\label{fig:teaser}
}

\maketitle

\begin{abstract}
Point cloud frame interpolation is a challenging task that involves accurate scene flow estimation across frames and maintaining the geometry structure.
Prevailing techniques often rely on pre-trained motion estimators or intensive testing-time optimization, resulting in compromised interpolation accuracy or prolonged inference.
This work presents FastPCI that introduces Pyramid Convolution-Transformer architecture for point cloud frame interpolation. Our hybrid Convolution-Transformer improves the local and long-range feature learning, while the pyramid network offers multilevel features and reduces the computation. In addition, FastPCI proposes a unique Dual-Direction Motion-Structure block for more accurate scene flow estimation. Our design is motivated by two facts: (1) accurate scene flow preserves 3D structure, and (2) point cloud at the previous timestep should be reconstructable using reverse motion from future timestep. 
Extensive experiments show that FastPCI significantly outperforms the state-of-the-art PointINet and NeuralPCI with notable gains (\eg 26.6\% and 18.3\% reduction in Chamfer Distance in KITTI), while being more than $10\times$ and $600\times$ faster, respectively. Code is available at \url{https://github.com/genuszty/FastPCI}.
\end{abstract}

\begin{figure}[t]
\centering
\includegraphics[width=1.0\columnwidth]{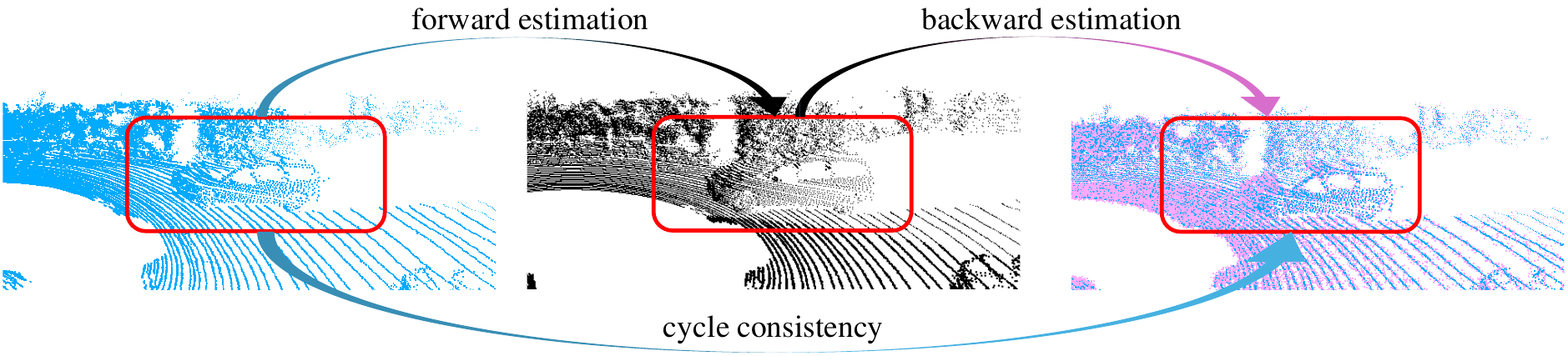}
\caption{(left) : frame $t-1$ ; (middle) : frame $t$ ; (right) : the overlapping original frame $t-1$ and the frame $t-1$ estimated from frame $t$. FastPCI produces structure-aware motion and performs dual-direction motion estimation motivated by two facts: 
(1) \textbf{structure consistency}: an accurate motion preserves structure of objects, \eg the car in red box; (2)  \textbf{cycle consistency}: point cloud at frame $t$ (right) is predicted from frame $t-1$ (left) by the estimated motion, meanwhile frame $t-1$  can be reconstructed from frame $t$ (middle) by applying the reverse estimated motion (right).
}
\label{fig:pullfig}
\end{figure}

\section{Introduction}
\label{sec:introduction}

LiDAR is widely used in autonomous driving, virtual/augmented reality, and robotics \cite{VR, 3Dtracking, humanpose}. 
However, due to hardware limitations, LiDAR usually captures low frame-rate scans that can lead to temporal discontinuities. Point Cloud Frame Interpolation (PCI) addresses this challenge by synthesizing intermediate frames, which has the wide application in enhancing the temporal resolution of the data and improving the efficacy of applications reliant on continuous environmental representation, such as real-time tracking and navigation \cite{motionforecasting,3Dtracking}.

Early work \cite{Plin_pseudo-lidar, Pseudo-lidar} relied on stereoscopic images to generate pseudo-LiDAR point cloud interpolation. Recent works \cite{PointPWC-Net, NSFP, PointINet, IDEA-Net, NeuralPCI} adopt neural networks to predict interpolated point clouds end-to-end in a data-driven manner. IDEA-Net \cite{IDEA-Net} refines the linear interpolation results by learning a one-to-one alignment matrix and modeling a trajectory compensation module. However, the one-to-one correspondence assumption in IDEA-Net limits its application on large outdoor point cloud datasets. 
PointINet \cite{PointINet} predicted the interpolated point cloud through learning fusion and refinement networks with a fixed pretrained motion estimator. However, it relies merely on pre-trained motion estimation and cares less about structure consistency, leading to noisy prediction as illustrated in \figlabel\ref{fig:teaser}.
Most recently, NeuralPCI \cite{NeuralPCI} adopted a dynamic neural radiance field \cite{NeRF, dnerf} and achieved state-of-the-art accuracy in PCI. Unfortunately, NeuralPCI requires testing-time per-scene optimization, which is rather expensive.

In contrast to pioneering PCI methods, our work is based on two facts, as illustrated in \figlabel\ref{fig:pullfig}: (1) \textit{structure consistency}: an accurate motion preserves the structure of objects; (2) \textit{cycle consistency}: point cloud at frame $t$ is predicted by the estimated motion $M_{t-1,t}$ from the frame $t-1$, and a reversed motion $M_{t,t-1}$ should reconstruct the point cloud in the frame $t-1$. 
Respecting these two consistencies, we propose a Dual-Direction Motion-Structure Transformer block. Given the point features from frame 0 and frame 1, we concatenate them in forward and backward order and use them for a bidirectional input in the block to encourage the model be aware of the motion direction. We perform a cross-attention between the bidirectional input to estimate the motion and structure jointly. 
We hierarchically stack this Motion-Structure block, and propose our Pyramid Motion-Structure Estimation Network. The pyramid structure provides multilevel features and reduces the computational burden. The usage of Transformer enhances the global reasoning ability due to its attention mechanism. We finally propose FastPCI consisting of a Pyramid Motion-Structure estimation network, a motion compensation network, and an interpolation refinement network for point cloud frame interpolation. Through extensive experiments in three large-scale automated driving datasets, we demonstrate significant improvements over the state-of-the-art in both accuracy and speed, as shown in \figlabel\ref{fig:teaser}.

\noindent\textbf{Contributions} of our work are listed as follows: 
\begin{itemize}[leftmargin=10pt,nosep] 

\item We present Dual-Direction Motion-Structure Transformer block. This block estimates motion in a structure-aware manner through the mixed information from forward and backward point features.

\item We propose FastPCI that adopts a Pyramid Convolution-Transformer architecture tailored for rapid and precise point cloud frame interpolation.

\item We propose to optimize FastPCI with reconstruction loss and additional pyramid loss and dual-direction losses. Comprehensive evaluations across diverse autonomous driving datasets demonstrate FastPCI's superiority over the state-of-the-art.
\end{itemize}

\section{Related Work}\label{sec:related}

\innersection{Point Cloud Frame Interpolation.}
Existing point cloud interpolation methods try to find point-to-point correspondences between two point cloud reference frames. They can be divided into two categories. One is to explicitly predict continuous interframe motion by estimating scene flow. PointINet \cite{PointINet} warps two input frames with bi-directional flows to the intermediate frame, then adaptively samples the two warped results and fuses them with attentive weights. This method is implemented based on pre-trained scene flow, focusing more on local motion. The effectiveness of frame interpolation of PointINet highly depends on the accuracy of the scene flow. The other is an implicit field estimation technique that requires testing-time optimization. NeuralPCI \cite{NeuralPCI} creates a spatial-temporal neural field to fit the given frames, and is able to output interpolated frames due to its continuous modeling. The fatal disadvantage of NeuralPCI is the prolonged inference time. 
In our work, we still train a feedforward point cloud frame interpolation network that runs efficiently in inference. We present a hybrid convolution-transformer network with a pyramid architecture. Our model incorporates a structure-aware motion estimation and a dual-direction cross-attention as inspired by the structure consistency and cycle consistency. We also provide explicit losses to enhance pyramid feature learning through hierarchical supervision and to enforce cycle consistency. 

\innersection{Video Frame Interpolation}
The goal of the video frame interpolation (VFI) task is that the predicted frame at any time should not only satisfy the continuity of the dynamic objects but also have the same smooth appearance as the input frame. Therefore, the existing VFI methods can be regarded as the process of capturing the motion between successive frames, or explicitly or implicitly mixing the appearance consistent with the input frame to synthesize the intermediate frame. Many VFI methods perform motion estimation and appearance extraction in a hybrid manner \cite{DAIN, RIFE, IFRNet, VFIFormer}, warp the input frame with the help of optical flow, and then mix the appearance information to enhance the detail and generate an intermediate frame. Due to the need to handle both motion and appearance, this approach requires well-designed and high-capacity extraction modules. There are also many ways to design separate modules to extract motion and appearance information \cite{CAS-VFI, BMBC, FLAVR, XVFI}, and the addition of additional modules often creates a high computational overhead. To deal with complex motions and reduce the complexity of the model, EMA-VFI \cite{EMA-VFI} proposes an interframe attention module, which is integrated into the hybrid architecture of CNN and transformer, and reuses attention mapping to extract appearance features and motion features, saving the computational overhead of a single model. These works inspire us to design a point cloud interpolation network that extracts both motion and structure-consistent information end-to-end. However, extending these approaches to unordered and unstructured point clouds remains a challenge.

\innersection{General Point Cloud Processing}
With the popularity of three-dimensional scanning devices such as lidar, laser, or RGB-D scanners, the capture of point clouds has become easier, and point cloud processing technology has gained great attention. Earlier work transformed point clouds into regular grids or 3D voxels, which are input into a typical convolutional architecture. However, this data representation transformation loses the point cloud structure and introduces quantization artifacts. The pioneering point-based work is PointNet \cite{PointNet}, which builds a framework capable of directly processing point cloud data using operators with permutation invariance, such as point-by-point MLP and pooling layers, to aggregate features across sets. PointNet++ \cite{PointNet++} compensates for the shortcomings of pointnet and achieves global and local detail feature extraction. PointConv \cite{PointConv}, DGCNN \cite{DGCNN} and Point transformer \cite{pointtransformer} are pioneering frameworks based on point convolution, graph convolution, and transformer, respectively. Followups \cite{DeepGCNs, ASSANet, qian2022pix4point, PointNeXt, KPConv} propose different designs to improve the efficacy or efficiency of point cloud learning. Our work is inspired by these 3D point cloud processing \cite{PointTransformerV2}. We introduce a hybrid Convolution-Transformer framework with a pyramid structure into the point cloud frame interpolation field, along with novel filed-specific designs to push the boundary for state-of-the-art point cloud frame interpolation. 
\section{Method} 
\label{sec:method}


In this section, we present FastPCI, a Pyramid Convolution-Transformer architecture for fast point cloud frame interpolation. 
FastPCI adopts Motion-Structure Transformer along with a Dual-Direction Cross-Attention to estimate structure-aware motion using bidirectional features. We also propose architecture-related loss functions to further improve performance and will elaborate next.

\begin{figure}[t]
\centering
\includegraphics[trim={0, 0, 0, 0}, clip, width=1.0\columnwidth]{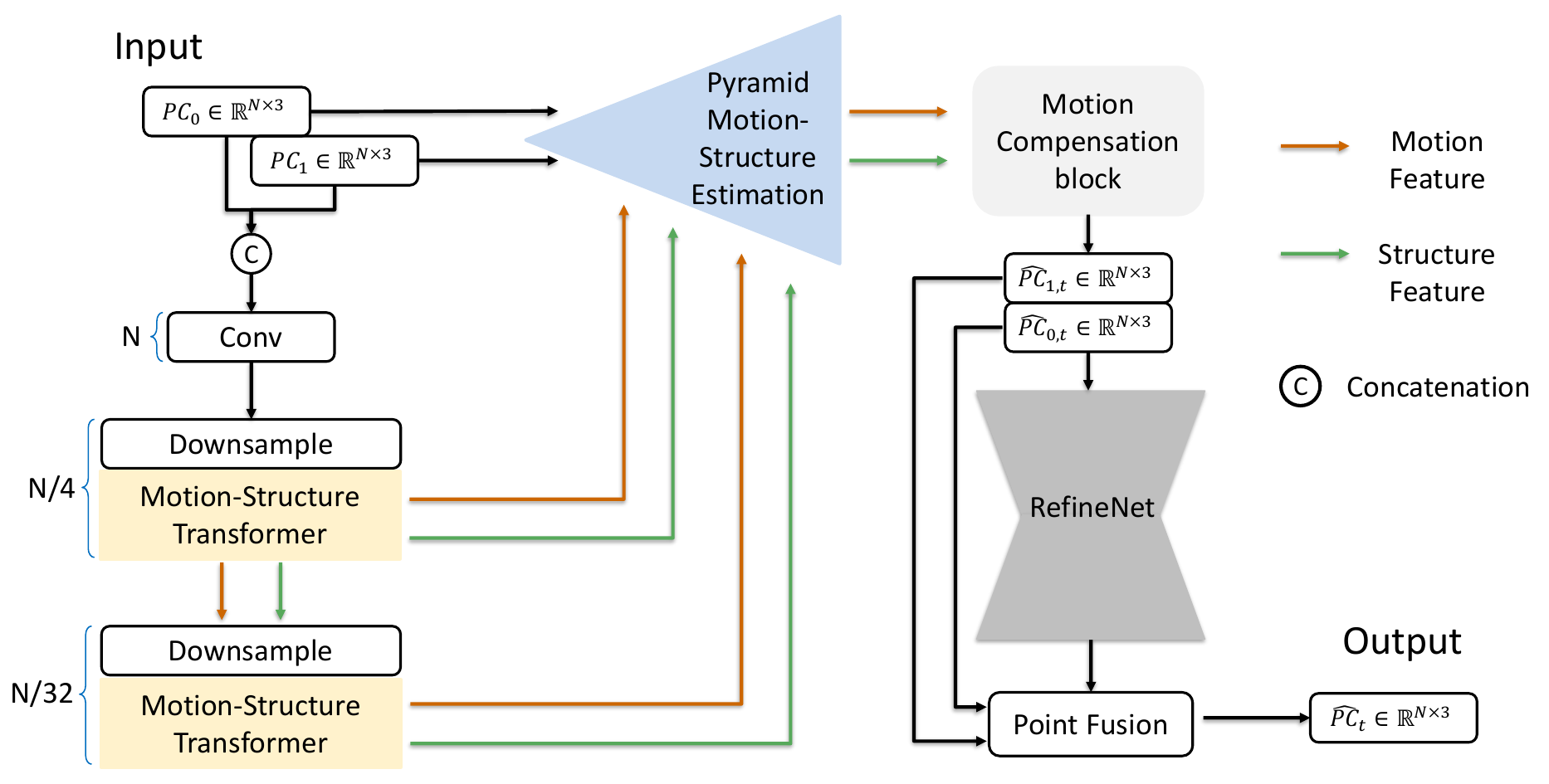}
\caption{\textbf{Overview of FastPCI pipeline.} Given two input frames $PC_0\in\mathbb{R}^{N \times 3}$ and $PC_1\in\mathbb{R}^{N \times 3}$, FastPCI estimates both motion and structure using a Pyramid Convolution-Transformer network. The estimated motion is used to warp the input frame to produce interpolated frames. RefineNet further refines the interpolated frames, and outputs the final frames from the fused forward and backward estimates. 
}
\label{fig:pipeline}
\vspace{-1.0em}
\end{figure}

\subsection{Overall Pipeline}
Our overall pipeline is illustrated in \figlabel\ref{fig:pipeline}. 
Given two consecutive point clouds with low temporal resolution $PC_{0}\in\mathbb{R}^{L \times 3}$ and $PC_{1}\in\mathbb{R}^{L \times 3}$ in the time step $t=0$ and $t=1$, the goal of FastPCI is to generate the point cloud frame $PC_{t}\in\mathbb{R}^{L \times 3}$ in arbitrary time step $t\in(0,1)$. 
Here $L$ denotes the number of points in the current point cloud.
FastPCI first estimates the motion feature $M \in \mathbb{R}^{L \times C}$ and structure feature $S \in \mathbb{R}^{L \times C}$ from the proposed Pyramid Convolution-Transformer Estimation network. 
The motion feature $M$ is the high-dimensional representation of the scene flow $SF$ that represents the displacement between the corresponding points of consecutive input frames. Throughout the paper, $M$ and $M_{0,1}$ represent forward motion from time $0$ to time $1$, and $M_{1,0}$ represent reverse motion from time $1$ to $0$. 
Then, we refine our motion estimation through a compensation block taking $M$ and $S$ as input and output the scene flow $SF$. A dual-direction estimation of interpolated point clouds $\widehat{PC}_{0,t}$ and $\widehat{PC}_{1,t}$ can be obtained by adopting forward and reverse motions from time steps $0$ and $1$, respectively. Here, $\widehat{PC}_{0,t}$ is the forward estimate of $\widehat{PC}_{t}$ from $PC_{0}$ and the backward estimate of $\widehat{PC}_{t}$ from $PC_{1}$.   
Then, a RefineNet is optimized to update the forward estimate $\widehat{PC}_{0,t}$. Finally, we follow the same fusion strategy as PointINet \cite{PointINet} to get the final output from the refined forward estimate and the backward estimate. The entire system is trained end-to-end with ground truth ${PC}_{t}$.

We highlight two unique designs in our pipeline which we elaborate on next:
(1) Pyramid Motion-Structure Network for structure-aware motion estimation. Keeping the structure of the front and back frames consistent is crucial for point cloud frame interpolation. This motivates our design of structure-aware motion estimation, which differs from the state-of-the-art considering only motion. Our network also employs Pyramid Convolution-Transformer for fast processing and multilevel feature learning. Our novel use of Transformer for point cloud frame interpolation also adds a global reasoning ability to the network and overcomes the limitation of local processing of point convolutions. 
(2) Motion-Structure Transformer Block that introduces dual-direction cross-attention for improved motion estimation by considering both forward and backward point features.

\subsection{Pyramid Motion-Structure Estimation}

Our Pyramid Motion-Structure Estimation Network takes two frames at time $0$ and $1$ as input and estimates the forward motion features and structure features from frame $0$ to frame $t$. The output of this network is fed into a Motion Compensation to get the scene flow, \ie motion in coordinate space.

Pyramid Motion-Structure Estimation consists of three pyramid stages. The first stage is built using a three-layer MLP (a mini-PointNet \cite{PointNet}). 
The second and third stages are built using a downsampling layer to reduce the initial spatial resolution $N$ by $4 \times$ and $32\times$ respectively. They are all followed by a cost layer and an SF Predictor (details in \textbf{Appendix}). Multilevel motion and structure features estimated by our Motion-Structure Transformer are fed into the SF Predictor of the second and third stages. At each stage, motion and structure features are captured from the point features $F_0^{l}$, $F_1^{l}$ by the Motion-Structure Transformer block. Here, $F$ and $l$ denote the point features and the $l^{th}$ stage.

\begin{figure}[t]
\centering
\includegraphics[trim={0, 0, 0, 0}, clip, width=1.0\columnwidth]{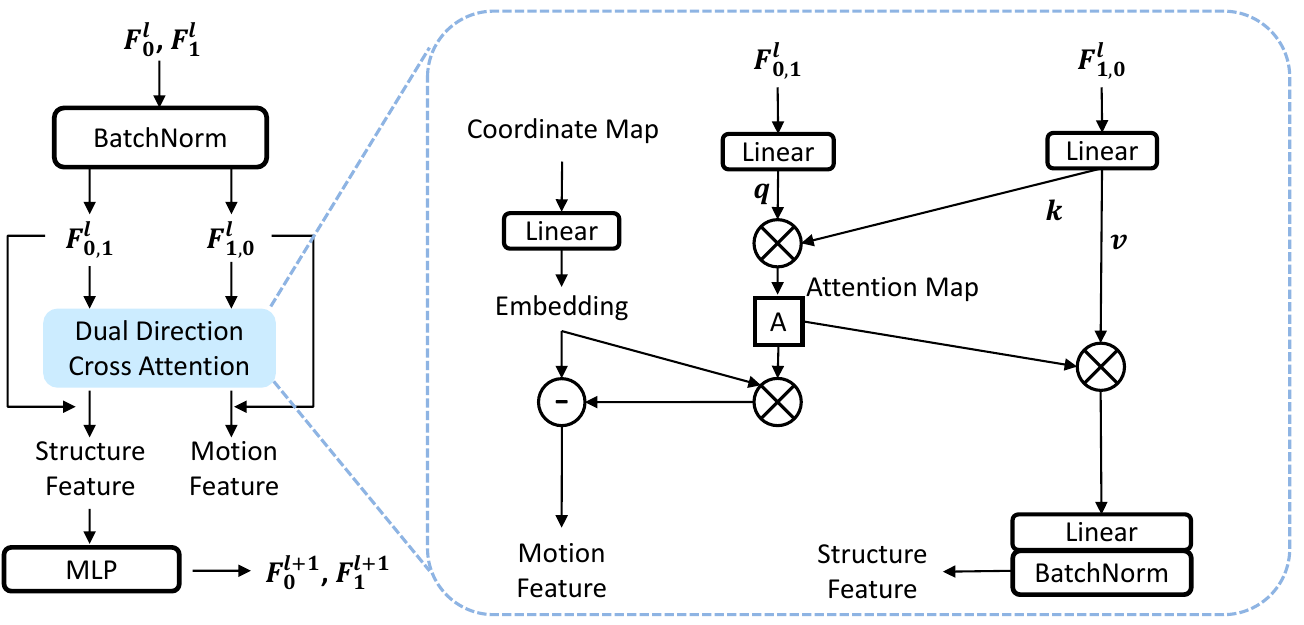}
\caption{\textbf{The illustration of Motion-Structure Transformer.} Where $\otimes$ and $\ominus $ denote matrix multiplication and element-wise subtraction, respectively. We highlight our Motion-Structure Transformer takes a bidirectional point features input, and perform a Dual-Direction Cross-Attention across forward and backward features. The structure and motion features are closely related to each other to learn a structure-aware motion. 
}
\label{fig:MSTransformer}
\end{figure}

\subsection{Motion-Structure Transformer}
As illustrated in \figlabel\ref{fig:MSTransformer}, our Motion-Structure Transformer estimates motion features and structure features from point features. 
In detail, given features $F_{0}^l\in\mathbb{R}^{L\times C}$ and $F_{1}^l\in\mathbb{R}^{L\times C}$ of the timestep $0$ and $1$ at stage $l$, we normalize the features, and stack them in forward and backward order, which gives a dual-direction features ${F}_{0,1}^l\in\mathbb{R}^{L \times C} = [{F}_{0}^l;{F}_{1}^l] $ and 
${F}_{1,0}^l\in\mathbb{R}^{L \times C} = [{F}_{1}^l;{F}_{0}^l] $, where $L, C, ;$ denote the number of points, channel size, and concatenation operation in the batch dimension, respectively. 
A subsequent Dual-Direction Cross Attention module is used to output forward motion and structure features by using the forward feature as query and the backward feature as key and value. 
Mathematically, query, key, and value in the attention is given by: 

\begin{equation}\label{eq1}
    q_{0,1}={F}_{0,1}^lW_{q}, \hfill k_{1,0}={F}_{1,0}^lW_{k},  \hfill v_{1,0}={F}_{1,0}^lW_{v}, 
\end{equation}
where $W_{q}$, $W_{k}$, $W_{v} \in\mathbb{R}^{C\times{\hat{C}}}$ are linear projection weights. Then we calculate the dual-direction cross-attention map and the structure features as follows:
\begin{align}\label{eq2}
    A=SoftM&ax(q_{0,1}k_{1,0}^\mathrm{T}/\sqrt{C}), \\
    S=&W_{S}(Av_{1,0}),
\end{align}
where $W_S$ represents linear projection for the structure features. The structure features are passed to a two-layer MLP to update points features $F_0^{l+1}$, $F_1^{l+1}$ for the next stage.

For motion estimation, we create a coordinate map $B\in\mathbb{R}^{L \times 3}$, where the value at each location indicates the unit position that expands uniformly along each direction and covers the entire point cloud frame. A higher dimensional embedding of $B_1$ is obtained in each cross-attention block by applying a single linear layer on $B$. The coordinate embeddings are multiplied by attention map $A$ to estimate the corresponding position of the point after the motion. Finally, the motion vector $M_{0,1}$ is estimated by applying another linear projection on the difference between the original $B_1$ and the wrapped $B_1$ by attention map:
\begin{equation}\label{eq3}
    M_{0,1}=W_{M}(A{B_1}-{B_1}), 
\end{equation}
where $W_M$ represents linear projection for the motion features. It is worth noting that under the assumption of local linear motion, the motion features and structure features obtained by the Motion-Structure Transformer module can be multiplied by $t$ to synthesize intermediate frames at any $t$ moment, where $t \in [0,1]$. 
We also highlight that the attention $A$ comes from the point features $F^l$, which are learned from the structure features. Therefore, the motion and structure features in our block are closely related. The motion features estimated from the Motion-Structure Transformer are structure-aware.

\subsection{Motion Compensation Block}
Inspired by video frame interpolation  \cite{VFIFormer}, we propose Motion Compensation Block to refine motion. 
In contrast to \cite{VFIFormer} and different from point cloud frame interpolation methods, we take the concatenation of motion and structure features as input.  Our pivotal consideration of both motion and structure enhances the capture of motion relationships and the preservation of structural integrity between frames.
Specifically, first, a three-layer MLP and sigmoid are leveraged to enhance nonlinear structure features, while input motion features are fed into one-dimensional convolution and LeakyReLU \cite{LeakyReLU} for enhancement.
Then, we concatenate the enhanced structure and motion features to a motion offset head built by two layers of convolution to obtain a motion correction term. The motion correction term is added back to the input motion to get the refined motion features. From the motion features, we use a scene flow prediction layer to map the motion features to $SF^l \in \mathbb{R}^{L\times3}$. 


Due to our hierarchical architecture, we can get a multilevel scene flow. A warping operation can be applied to obtain an interpolation frame at each timestep $t \in [0, t]$ for each stage by:
\begin{equation}\label{eq5}
    \widehat{PC}_{0,t}^l = \widehat{PC}_0^l + SF^l * t, \quad 
    \widehat{PC}_{1,t}^l = \widehat{PC}_1^l + SF^l * (1-t). 
\end{equation}

Specifically, the final clean scene flow is used to predict the final output. The scene flows in each other level are used for supervised pyramid prediction.

\subsection{RefineNet}
We use an additional RefineNet to refine the forward estimate $\widehat{PC}_{0,t}$ since it is more sensitive to the final results. We simply leverage a three-stage U-Net \cite{U-Net} to encode and decode the input points. After each upsampling and downsampling layer, we use a Point Transformer block \cite{pointtransformer} to enhance features. The output of the UNet is concatenated with our backward estimate to yield a fused point cloud. We then closely follow PointINet \cite{PointINet} to get the final interpolated point cloud from the fused estimates. Refer to their paper for the details on estimating the final output from the fused estimation.

\subsection{Objective Function}
Following previous work \cite{NeuralPCI, PointINet}, the entire system can be end-to-end trained by Chamfer Distance (CD) between the predicted point cloud frame $\widehat{PC}_{t}\in\mathbb{R}^{N \times 3}$ at time $t$ and the Ground Truth $PC_{t}\in\mathbb{R}^{N \times 3}$, as follows:
\begin{equation}\label{eq:reconstruct}
    L_{intp} = \frac{1}{N} \sum_{\hat{x}^{i}\in\widehat{PC}_t} \underset{x^{j}\in{PC}_t}{min} {\left \| \hat{x}^{i}-x^{j} \right \|}_{2} + \frac{1}{N} \sum_{x^{j}\in PC_t} \underset{\hat{x}^{i}\in\widehat{PC}_t}{min} {\left \| \hat{x}^{i}-x^{j} \right \|}_{2},
\end{equation}

In addition to this loss, we find that the following two regularizations can significantly improve the performance. 
First is a \textbf{half-way cycle consistency }that forces both of the forward estimate $\widehat{PC}_{0,t}$ and the backward estimate $\widehat{PC}_{1,t}$ to be close to the Ground Truth. Note we do not directly enforce a cycle loss, \ie reconstructing frame 0 from frame 1 using reverse motion, since it might encourage the network to learn an identity mapping. Mathematically, the half-way cycle loss encourages cycle consistency as well. This half-way cycle loss is given by the sum of two CD losses: 
\begin{align}\label{eq:cycle}
    L_{{CD}_1} = CD(\widehat{PC}_{0,t}, PC_t), \\
    L_{{CD}_2} = CD(\widehat{PC}_{1,t}, PC_t),
\end{align}

The second regularization is the \textbf{pyramid reconstruction loss}. Thanks to the hierarchical architecture, 
FastPCI offers predicted frames of different sizes. We thus propose the following multiscale loss $L_{ms}$ following the pyramid structure as shown next:
\begin{equation}\label{eq10}
    L_{ms} = \sum \alpha_{l}(CD(\widehat{PC}_{t}^{l}, PC_t^{l})),
\end{equation}
where $PC_t^{l}$ is obtained through furthest point sampling from the Ground Truth. $\alpha_l$ denotes the loss weight for pyramid level $l$. Throughout the experiments, we use $\alpha_0=0.05$, $\alpha_1=0.1$, $\alpha_2=0.2$. We empirically find that as long as $\alpha$ is set to $(0.025-0.25)$, the final performance is similar. 

Overall, FastPCI is optimized by the sum of all losses and regularization:
\begin{equation}\label{eq:loss}
    L = L_{intp} + L_{{CD}_1} + L_{{CD}_2} + {L_{ms}}.
\end{equation}

\section{Experiments}\label{sec:exp}
\subsection{Experimental Setup}
\textbf{Datasets.} 
The proposed FastPCI is assessed on three large outdoor LiDAR datasets, namely KITTI odometry \cite{KITTI}, Argoverse 2 sensor \cite{Argoverse2}, and Nuscenes \cite{Nuscenes}. The KITTI odometry and Argoverse 2 sensor data are point cloud frames collected at a frame rate of 10Hz, whereas the Nuscenes data are collected at 20Hz. To align datasets, the Nuscenes data are first downsampled to 10Hz. We follow NeuralPCI \cite{NeuralPCI} to split the training and testing sets. 
In detail, KITTI odometry dataset contains 11 annotated LiDAR point cloud sequences in total, where the first 7 sequences are used as training and the left is used as testing. Argoverse 2 sensor dataset is composed of 1,000 scenarios with 150 LiDAR sweeps per scenario on average, while Nuscenes dataset also consists of 1,000 driving scenes with about 400 LiDAR frames for each scene. For these two datasets, we use the first 700 scenes to train, and 850-1000 scenes to test. For all datasets, the input spatial resolution is 8192 points, and the temporal resolution is 2Hz. For evaluation, we estimate 3 intermediate frames per two frames. 


\noindent\textbf{Implementation Details.} 
We use PyTorch \cite{Pytorch} to implement FastPCI. We train FastPCI in all three training sets using the same parameters on a NVIDIA GeForce RTX 3090 GPU. FastPCI is optimized by Adam \cite{Adam} with a batch size of 4, a weight decay of $10^{-4}$, and an initial learning rate of $10^{-3}$ that is reduced by half every 80 epochs. 
We train KITTI odometry, Argoverse 2 sensor, and Nuscenes datasets by 100, 200, and 200 epochs, respectively.

\noindent\textbf{Evaluation Metrics.} 
We adopt CD and Earth Mover’s Distance (EMD) as quantitative evaluation metrics following \cite{PointINet, NeuralPCI}. EMD measures the minimum average distance required to move data from a point cloud to another point cloud.
Given two point clouds $\widehat{PC}_{t}\in\mathbb{R}^{N \times 3}$ and $PC_{t}\in\mathbb{R}^{N \times 3}$, EMD can be calculated as follows, where $\phi : \widehat{PC}_t \to PC_t$ is the bijection set:
\begin{equation}\label{eq12}
    L_{EMD} = \underset{\phi:\widehat{PC}_t \to PC_t }{min} \frac{1}{N} \sum_{\hat{x}\in\widehat{PC}_t} {\left \| \hat{x}-\phi(\hat{x}) \right \|}_{2}
\end{equation}

\begin{table*}[t]
    \centering
    \caption{\textbf{Quantitative comparison with the state-of-the-art on KITTI odometry, Argoverse 2 sensor, and Nuscenes}. \emph{Frame-1}, \emph{Frame-2} and \emph{Frame-3} refer to the three uniformly intermediate frames to be interpolated between the two input frames. Average denotes the average result on these three frames. \textbf{Bold} represents the best performance across different methods. }
    \label{tab:sota}
    \resizebox{1.0\textwidth}{!}{
    \begin{tabular}{l|c|cccccc|ccc}
    \toprule
    \multirow{2}{*}{\textbf{Dataset}} & \multirow{2}{*}{\textbf{Methods}} & \multicolumn{2}{c}{\textbf{Frame-1}} & \multicolumn{2}{c}{\textbf{Frame-2}} & \multicolumn{2}{c}{\textbf{Frame-3}} & \multicolumn{2}{|c}{\textbf{Average}} \\ 
    & & \textbf{CD} & \textbf{EMD} & \textbf{CD} & \textbf{EMD} & \textbf{CD} & \textbf{EMD} & \textbf{CD$\downarrow$} & \textbf{EMD$\downarrow$} \\ \midrule
    \multirow{5}{*}{KITTI odometry}&{PointPWC-Net \cite{PointPWC-Net}}&{0.64}&{71.14}&{0.80}&{91.91}&{0.91}&{60.35}&{0.78}&{74.46} \\  
    &{NSFP \cite{NSFP}}&{0.58}&{70.53}&{0.68}&{84.76}&{1.95}&{115.42}&{1.07}&{90.24} \\
    &{PointINet \cite{PointINet}}&{0.72}&{55.25}&{0.82}&{77.87}&{0.83}&{73.74}&{0.79}&{69.19} \\
    &{NeuralPCI \cite{NeuralPCI}}&{0.64}&{52.61}&{0.85}&{62.73}&{0.64}&{52.15}&{0.71}&{55.83} \\
    &{\textbf{FastPCI(our)}}&{\textbf{0.54}}&{\textbf{51.23}}&{\textbf{0.61}}&{\textbf{59.84}}&{\textbf{0.58}}&{\textbf{50.27}}&{\textbf{0.58}}&{\textbf{53.78}} \\ \midrule
    
    \multirow{5}{*}{Argoverse 2 sensor}&{PointPWC-Net \cite{PointPWC-Net}}&{0.90}&{56.44} &{1.07}&{79.86} &{1.26}&{65.32} &{1.07}&{67.20} \\
    &{NSFP \cite{NSFP}}&{0.72}&{62.30}&{0.85}&{73.89}&{2.14}&{98.99}&{1.24}&{78.40} \\
    &{PointINet \cite{PointINet}}&{0.83}&{57.89}&{1.25}&{67.73}&{1.06}&{62.97}&{1.05}&{62.86} \\
    &{NeuralPCI \cite{NeuralPCI}}&{0.68}&{55.03}&{0.88}&{65.93}&{\textbf{0.69}}&{55.30}&{0.75}&{58.75} \\
    &{\textbf{FastPCI(our)}}&{\textbf{0.68}}&{\textbf{54.99}}&{\textbf{0.77}}&{\textbf{64.29}}&{0.74}&{\textbf{54.59}}&{\textbf{0.73}}&{\textbf{57.95}} \\ \midrule
    
    \multirow{5}{*}{Nuscenes}&{PointPWC-Net \cite{PointPWC-Net}}&{1.52}&{172.31}&{1.39}&{224.64}&{2.16}&{181.39}&{1.69}&{192.78} \\
    &{NSFP \cite{NSFP}}&{1.10}&{173.03}&{1.33}&{212.32}&{4.37}&{319.56}&{2.27}&{234.97} \\
    &{PointINet \cite{PointINet}}&{1.48}&{183.03}&{1.67}&{\textbf{202.10}}&{1.50}&{186.51}&{1.55}&{190.54} \\
    &{NeuralPCI \cite{NeuralPCI}}& {\textbf{1.00}}& {163.10}& {1.37}& {205.24}&{1.06}&{168.98}& {1.15}& {179.11} \\
    &{\textbf{FastPCI(our)}}&{1.02}&{\textbf{162.78}}&{\textbf{1.28}}&{205.75}&{\textbf{1.03}}&{\textbf{152.39}}&{\textbf{1.11}}&{\textbf{173.64}} \\ \midrule
    \end{tabular}
    }
    \vspace{-1.0em}
\end{table*}


\subsection{Evaluation of Point Cloud Interpolation} 
To demonstrate the performance of FastPCI, we compare our method to previous SOTA methods, namely PointINet \cite{PointINet} and NeuralPCI \cite{NeuralPCI}, as well as the pioneering work \cite{PointPWC-Net, NSFP}. 
Due to the different preprocessing and data split across methods, it is not fair to compare with original results directly. Therefore, in our work, we use the same data splitting and preprocessing as in the most recent work and the state-of-the-art NeuralPCI. We report the original results of NeuralPCI and retrain other methods using the official implementations and the NeuralPCI's setups in all datasets.

\begin{figure}[t]
\centering
\includegraphics[trim={0, 0, 0, 0}, clip, width=1.0\columnwidth]{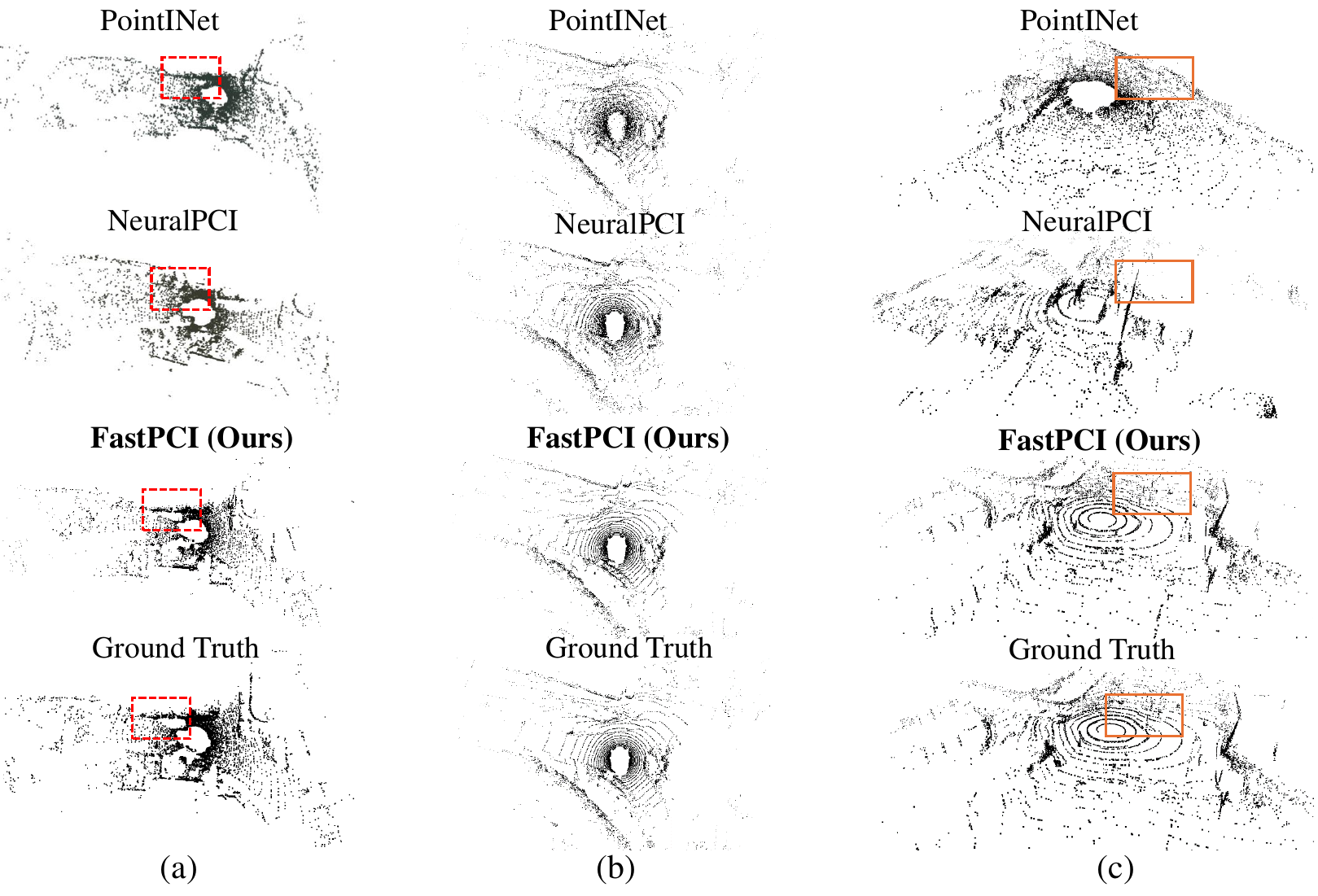}
\caption{\textbf{Qualitative comparisons with the state-of-the-art on KITTI odometry, Argoverse 2 sensor, and Nuscenes dataset.}  Columns (a)-(c) represent the results on three datasets, respectively. Each row represents a different method. Our FastPCI ($3^{rd}$ row) yields the best qualitative results compared to the state-of-the-art PointINet ($1^{st}$ row) and NeuralPCI ($2^{nd}$ row). 
}
\label{fig:visual-results}
\end{figure}

\noindent\textbf{Results on KITTI Odometry Dataset.} rows 3-7 of \tbllabel\ref{tab:sota} show the results on the KITTI odometry dataset, where FastPCI achieves the best results on all frames and in terms of all metrics. Specially, our approach reduces the EMD error by 2.89 and the CD error by 0.21 in the frame-2 compared to the state-of-the-art NeuralPCI \cite{NeuralPCI}, ultimately reducing 2.05 and 0.13 in the overall EMD and CD metrics results, respectively.
\figlabel\ref{fig:visual-results} (a) shows the visual comparisons among different methods on the KITTI odometry dataset. Highlighted in the red closeup, our FastPCI yields the best visual quality, producing the sharpest edge for vehicles with much less noise.

\noindent\textbf{Results on Argoverse 2 Sensor Dataset.} The quantitative comparison results are shown in rows 8-12 of \tbllabel\ref{tab:sota}, where FastPCI again achieves the best performance on nearly all frames across metrics. \figlabel\ref{fig:visual-results} (b) shows the qualitative experimental results of interpolated frames on the Argoverse 2 sensor dataset. Overall, the results of our method are closer to ground truth. Refer to the road and the boundary for differences.

\noindent\textbf{Results on Nuscences Dataset.} Rows 13-17 in \tbllabel\ref{tab:sota} display the quantitative results for Nuscenes. FastPCI has achieved the top performance on the overall index. Our method is slightly worse than the NeuralPCI \cite{NeuralPCI} method on the CD metric in frame-1 and the EMD metric in frame-2, but exceeds the NeuralPCI method by 5.47 and 0.04 on the overall EMD and CD metric. In the qualitative experiment, we present the frame interpolation visualization results of two advanced PCI methods and the proposed FastPCI on the Nuscenes dataset in \figlabel\ref{fig:visual-results} (c). From the contents of the yellow box, it is not difficult to find that compared with other methods, our method can insert clearer human outline lines.

\noindent\textbf{Runtime Comparison.} 
Fast point cloud frame interpolation can facilitate downstream applications in autonomous driving scenarios (such as target detection, target tracking, etc.). Therefore, we give the average inference time of a point cloud frame interpolated at any time in \tbllabel\ref{tab:time}. Compared with the current state-of-the-art point cloud frame interpolation methods, FastPCI shows the fastest inference times, being over 10 times faster than PointINet and 600 times faster than NeuralPCI, due to its Pyramid Convolutional-Transformer architecture.

\begin{table*}[t]
    \centering
    \caption{\textbf{Average inference time Comparison}. Time is measured per frame with 8192 points on a NVIDIA RTX 3090 GPU.}
    \label{tab:time}
    {\begin{tabular}{l|cccc}
    \toprule   
    \textbf{Methods} & \textbf{PointINet \cite{PointINet}} & \textbf{NeuralPCI \cite{NeuralPCI}} & \textbf{FastPCI(our)} \\ \midrule
    {Runtime} & {1458ms} & {60822ms} & \textbf{109ms} \\ \midrule
    \end{tabular}
    }
\end{table*}
\begin{table}[ht]
  \centering
  \begin{minipage}[c]{0.53\textwidth}
  \makeatletter\def\@captype{table}
    \caption{\textbf{Ablate architecture designs} on all three datasets. We show that our structure-aware motion estimation, the hybrid Convolution-Transformer architecture, the Dual-Direction Cross-Attention, the Motion Compensation block, and the RefineNet lead to improved performance in all datasets.}
    \label{table:ablation}
    \resizebox{1.0\textwidth}{!}{
    \begin{tabular}{p{1.4cm}|c|cc}
    \toprule
    \multirow{2}{*}{\begin{minipage}{1.4cm}\flushleft\textbf{Dataset}\end{minipage}} & \multirow{2}{*}{\textbf{Component}} & \multicolumn{2}{c}{\textbf{Metrics}} \\ 
    && \textbf{CD$\downarrow$} & \textbf{EMD$\downarrow$} \\ \midrule
    \multirow{6}{*}{\begin{minipage}{1.4cm}\flushleft KITTI odometry\end{minipage}}&{w/o structure branch}&{0.79}&{67.60} \\
    &{Transformer $\rightarrow$ convolution}&{0.61}&{56.38} \\
    &{w/o Dual Direction}&{0.60}&{56.34} \\
    &{w/o Motion Compensation}&{0.60}&{55.29} \\
    &{w/o RefineNet}& {0.60}& {54.53} \\ 
    &{\textbf{Baseline}}& \textbf{0.58}&\textbf{53.78} \\ \midrule
    \multirow{6}{*}{\begin{minipage}{1.4cm}\flushleft Argoverse 2 sensor\end{minipage}}&{w/o structure branch}&{0.85}&{62.23} \\
    &{Transformer $\rightarrow$ convolution}&{0.77}&{58.16} \\
    &{w/o Dual Direction}&{0.75}&{58.31} \\
    &{w/o Motion Compensation}&{0.75}&{58.14} \\
    &{w/o RefineNet}& {0.74}& {58.17} \\ 
    &{\textbf{Baseline}}& \textbf{0.73}&\textbf{57.95} \\ \midrule
    \multirow{6}{*}{\begin{minipage}{1.4cm}\flushleft Nuscenes\end{minipage}}&{w/o structure branch}&{1.64}&{193.95} \\
    &{Transformer $\rightarrow$ convolution}&{1.43}&{189.95} \\
    &{w/o Dual Direction}&{1.23}&{181.58} \\
    &{w/o Motion Compensation}&{1.20}&{179.35} \\
    &{w/o RefineNet}& {1.12}& {175.66} \\
    &{\textbf{Baseline}}& \textbf{1.11} & \textbf{173.64} \\ \midrule
    \end{tabular}
    }
  \end{minipage}
  \begin{minipage}[c]{0.4\textwidth}
  \makeatletter\def\@captype{table}
    \caption{\textbf{Ablate loss functions}. The proposed dual-direction reconstruction losses $L_{CD_{1}}$, $L_{CD_{2}}$, and the pyramid loss $L_{ms}$ improve the performance on all benchmarks. The dual-direction reconstruction losses are more crucial. 
    }
    \label{table:loss}
    \resizebox{1.0\textwidth}{!}{
    \begin{tabular}{p{1.4cm}|ccc|cc}
    \toprule
    \multirow{2}{*}{\begin{minipage}{1.4cm}\flushleft\textbf{Dataset}\end{minipage}} & \multicolumn{3}{c}{\textbf{Loss Item}} & \multicolumn{2}{c}{\textbf{Metrics}} \\ 
    & \textbf{$L_{CD_{1}}$} & \textbf{$L_{CD_{2}}$} &  \textbf{$L_{ms}$} & \textbf{CD$\downarrow$} & \textbf{EMD$\downarrow$} \\ \midrule
    \multirow{5}{*}{\begin{minipage}{1.4cm}\flushleft KITTI odometry\end{minipage}}&{$\surd$ }&{$\surd$}&{}&{0.73}&{60.99} \\
    &{$\surd$} &{} &{$\surd$} & {0.59}& {55.54} \\
    &{} &{$\surd$} &{$\surd$}&{0.74}&{58.47}  \\
    &{} &{} &{} &{0.78}&{67.59} \\
    &{$\surd$} &{$\surd$} &{$\surd$} &\textbf{0.58}&\textbf{53.78} \\ \midrule
    \multirow{5}{*}{\begin{minipage}{1.4cm}\flushleft Argoverse 2 sensor\end{minipage}} &{$\surd$} &{$\surd$} &{} & {0.80}&{59.35} \\
    &{$\surd$} &{} &{$\surd$} & {0.75}& {59.92} \\
    &{} &{$\surd$} &{$\surd$} &{0.83}&{60.57} \\
    &{} &{} &{} &{0.92}&{61.90} \\ 
    &{$\surd$} &{$\surd$} &{$\surd$} &\textbf{0.73}&\textbf{57.95} \\ \midrule
    \multirow{5}{*}{\begin{minipage}{1.4cm}\flushleft Nuscenes\end{minipage}} &{$\surd$}  &{$\surd$}  &{} &{1.37}&{184.28} \\
    &{$\surd$} &{} &{$\surd$}& {1.23}& {179.80} \\
    &{} &{$\surd$} &{$\surd$} &{1.56}&{185.15} \\ 
    &{} &{} &{} &{1.57}&{189.63} \\ 
    &{$\surd$} &{$\surd$} &{$\surd$} &\textbf{1.11}&\textbf{173.64} \\ \midrule
    \end{tabular}
    }
  \end{minipage}
\end{table}

\subsection{Ablation Study}

We contribute in both architecture design and loss functions as follows:
(1) Structure-aware motion Estimation; 
(2) Dual-Direction Estimation;
(3) Hybrid Convolution-Transformer architecture;
(4) Motion Compensation block;
(5) RefineNet;
(6) Cycle consistency loss;
(7) Multiscale loss.
We ablation each contribution through the following ablation study.

\subsection{Ablate Architecture Designs}

\noindent\textbf{Structure-Aware Motion Estimation} aims to keep the structure of the front and back frames consistent, which is crucial for the quality of the frame interpolation. To verify the effectiveness of our compound motion-structure learning, we remove all the structural branches and only keep the motion feature extraction in our network. As shown in \tbllabel\ref{table:ablation}, without a structure-aware design, a significant increase in the distances of CD and EMD can be observed. This experiment shows the importance of the Motion-Structure joint learning. 

\noindent\textbf{Dual-Direction Estimation} is designed to encourage the interaction of information between forward and backward frames. 
To show its effectiveness, we replace all reverse input features $F_{1,0}^l$ in \figlabel\ref{fig:MSTransformer} with forward features $F_{0,1}^l$. In other words, we use standard self-attention on forward features to substitute the original Dual-Direction Cross Attention. Shown in \tbllabel\ref{table:ablation}, without Dual Direction leads to a clearly worse performance, despite the existing cycle loss.

\noindent\textbf{Hybrid Convolution-Transformer} utilizes convolution to extract local features and Transformer for global reasoning ability.  We also ablate replacing the Transformer with regular convolutions. Due to the lack of Transformer, we do not perform cross attention anymore, and only use pure convolutions (two-layer mini-PointNet) to learn both motion and structure.
As shown in \tbllabel\ref{table:ablation} w/o Transformer,  significantly worse results are reported. These experiments point out the importance of Transformer in our design.

\noindent\textbf{Motion Compensation} block refines the estimated motion from motion and structure features. Here, we remove this block in \tbllabel\ref{table:ablation}, and see a slightly worse performance. This module marginally improves the performance, which is not the key component in FastPCI.

\noindent\textbf{RefineNet} is used to refine the interpolated point cloud. Similar to the Motion Compensation block, we remove the RefineNet and use the wrapped forward and backward estimates to fuse and get final results directly. \tbllabel\ref{table:ablation} w/o RefineNet shows a marginal performance drop, which indicates that the use of RefineNet can enhance results but the crucial part is Motion Compensation.



\subsection{Ablate Loss Functions}

\noindent\textbf{Half-way Cycle Consistency} aims to ensure the accuracy of forward and backward estimated motion. From \tbllabel\ref{table:loss}, one can observe a significant drop when removing $L_{CD_1}$, showing the importance of supervising the coarse estimation of frame interpolation.  The addition of backward warp loss $L_{CD_2}$ further improves cycle consistency, and thus improves the performance.

\noindent\textbf{Multiscale loss} provides hierarchical supervision for the interpolated frames estimated from the pyramid features. We empirically find that the multiscale loss contributes nontrivially to the performance of FastPCI. Along with the pyramid architecture, the hierarchical design in the network and the loss function are the most important designs in FastPCI and significantly boost performance. See \supp for the ablation study in the pyramid architecture for details. 

\section{Limitation}\label{sec:limitation}

While FastPCI has made great strides in point cloud frame interpolation, there are still some limitations. The effectiveness of this method under different environmental conditions and its robustness to outliers and occlusion need to be further studied. Furthermore, the adaptability of FastPCI to real-world and real-time deployment remains to be tested and explored.

\section{Conclusion}\label{sec:conclusion}
This paper introduces FastPCI, a pioneering Pyramid Convolution-Transformer hybrid architecture designed for rapid and accurate point cloud frame interpolation. 
The introduction of Transformer and the hierarchical architecture bring long-range information extraction capability to our system, while maintaining efficiency. 
We also observe the importance of 
structure consistency and cycle consistency to the task of point cloud frame interpolation. To respect both of them, our FastPCI proposes a unique Dual-Direction Motion-Structure estimation, to mix information between forward and backward estimates, as well as between motion and structure features. 
We also present two losses closely related to our architecture design. The first is the half-way cycle consistency loss that encourages the network to learn a cycle-consistent motion estimation. The second is a multilevel reconstruction loss that allows middle-stage supervision using lower-resolution ground truth, thanks to our hierarchical architecture design.  
Overall, FastPCI establishes a new benchmark in point cloud frame interpolation, surpassing the state-of-the-art methods PointINet and NeuralPCI with notable margins while being much more efficient.

\clearpage  

%
%
\bibliographystyle{splncs04}
\bibliography{egbib}
\end{document}









\title{FastPCI: Motion-Structure Guided Fast Point Cloud Frame Interpolation \\
--- Appendix ---}

\titlerunning{FastPCI}
\author{Tianyu Zhang\inst{1}  \and
Guocheng Qian\inst{2}\orcidlink{0000-0002-2935-8570} \and
Jin Xie\inst{3\dag} \and Jian Yang\inst{1}
\thanks{$\dag$ Corresponding author.}}

\authorrunning{T.~Zhang et al.}
\institute{Nankai University, Tianjin, China \and
Snap Research \and 
State Key Laboratory for Novel Software Technology, Nanjing University, Nanjing, China
School of Intelligence Science and Technology, Nanjing University, Suzhou, China \\
\email{tianyu.zhang@mail.nankai.edu.cn}, \email{guocheng.qian@kaust.edu.sa} \\
\email{\{csjxie, csjyang\}@njust.edu.cn} }

\maketitle

\renewcommand{\thesection}{\Alph{section}}
\renewcommand{\thetable}{S\arabic{table}}
\renewcommand{\thefigure}{S\arabic{figure}}
\setcounter{section}{0}

Here, we provide additional information to supplement the main paper. Firstly, we provide more details on the network structures of the Pyramid Motion-Structure Estimation module, the Motion Compensation block, and RefineNet in \S\ref{sec:supp_method}. In \S\ref{sec:supp_exp}, we carry out additional ablation studies on the Dual-Direction Cross Attention. 
Moreover, we provide additional qualitative results in \S\ref{sec:supp_visual}.

\begin{figure}[t]
\centering
\includegraphics[trim={0, 0, 0, 0}, clip, width=1.0\columnwidth]{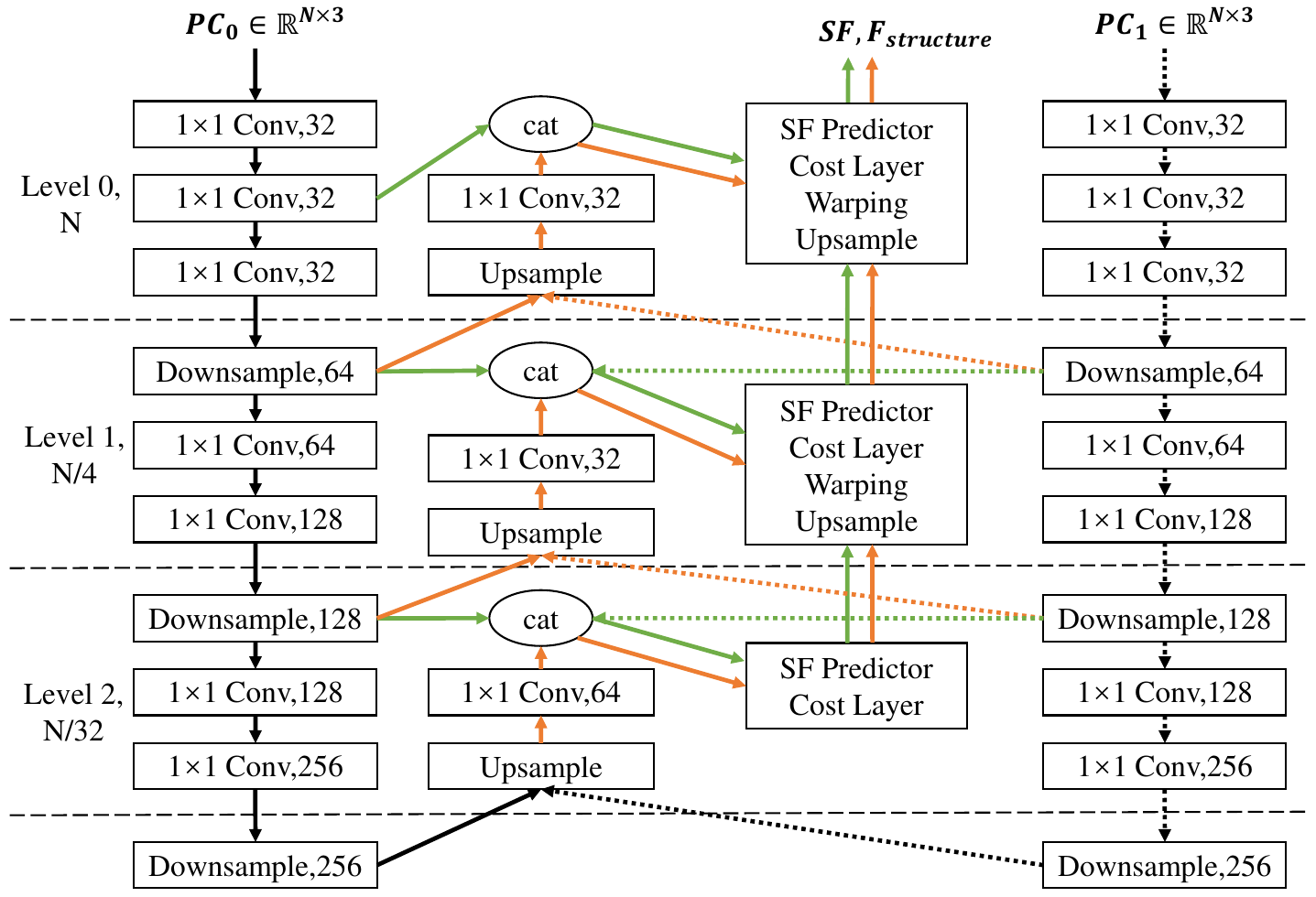}
\caption{\textbf{The illustration of Pyramid Motion Structure Estimation.} The input first point cloud and the second point cloud are encoded using the same network with shared weights. For each point cloud, we use Downsampling \cite{PointConv} to convolve and downsample by a factor of 4. The 1 × 1Convs are
used to increase the representation power and efficiency. The \textit{$cat$} represents a concatenation operation, and we use Cost Layer \cite{PointPWC-Net} and SF Predictor \cite{PointPWC-Net} to calculate cost volumes to predict motion and structure features. 
The cost layer find nearest neighbors between frames for accurate
motion estimation, which we  adopt from \cite{PointPWC-Net} and we omit this implementation detail in main paper since it is not our focus.
}
\label{fig:figureS1}
\end{figure}

\begin{figure}[t]
\centering
\includegraphics[trim={0, 0, 0, 0}, clip, width=0.5\columnwidth]{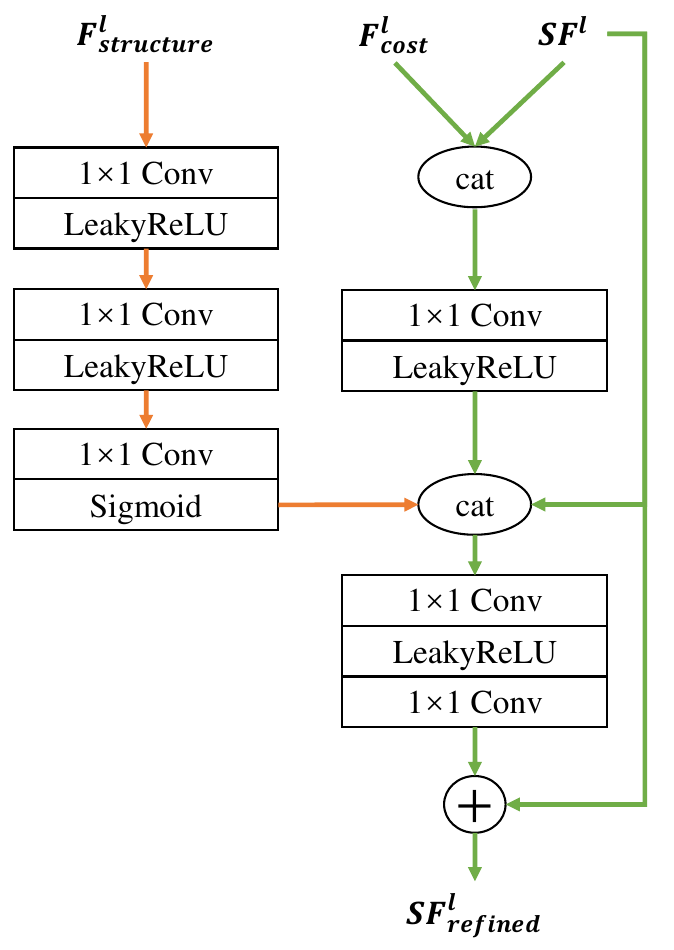}
\caption{\textbf{The illustration of Motion Compensation block.} 
Where \textbf{$F_{structure}^{l}$}, \textbf{$F_{cost}^{l} $}, and \textbf{${SF}^{l}$ } denote the structure feature, cost volume, and predicted scene flow from the pyramid level $l$, respectively. The \textit{$cat$ } is the concatenation operation, and $\oplus$ is element-wise addition. We output the refined scene flow \textbf{${{SF}_{refined}}^{l}$ } through the Motion Compensation block.
}
\label{fig:figureS2}
\end{figure}

\begin{figure}[t]
\centering
\includegraphics[trim={0, 0, 0, 0}, clip, width=1.0\columnwidth]{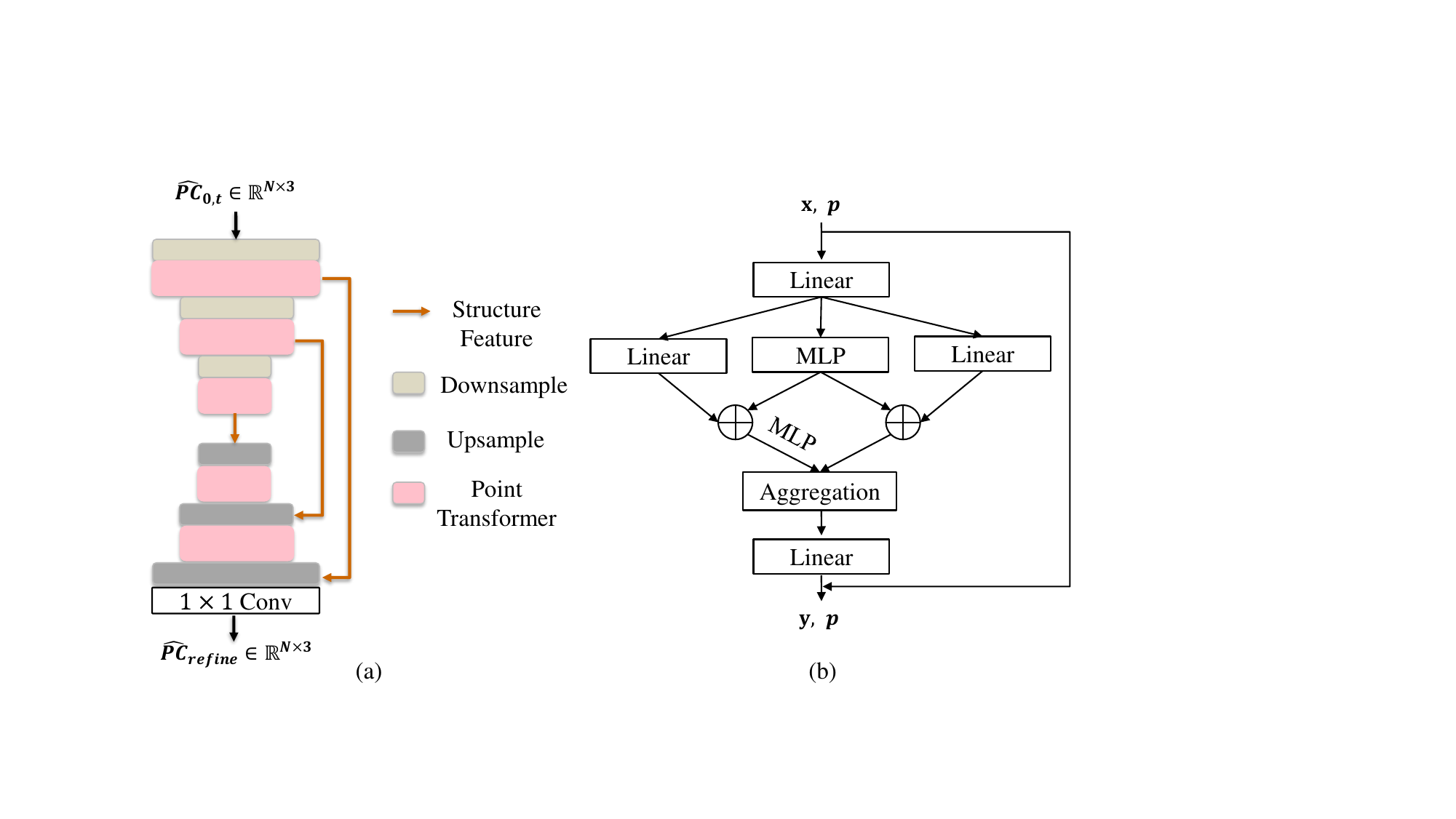}
\caption{\textbf{The illustration of RefineNet.} (a) Overview of RefineNet; (b) Mechanism of Point Transformer. Where $\oplus$ is element-wise addition. RefineNet is a three-layer U-Net-style module that corrects points in the point cloud at time $t$ after warp and outputs the refined warp point cloud. Point Transformer we adopt from \cite{pointtransformer} is used to converge structure features of different resolutions and to learn small shifts. 
}
\label{fig:figureS3}
\end{figure}

\begin{figure}[t]
\centering
\includegraphics[page=1,width=1.0\columnwidth]{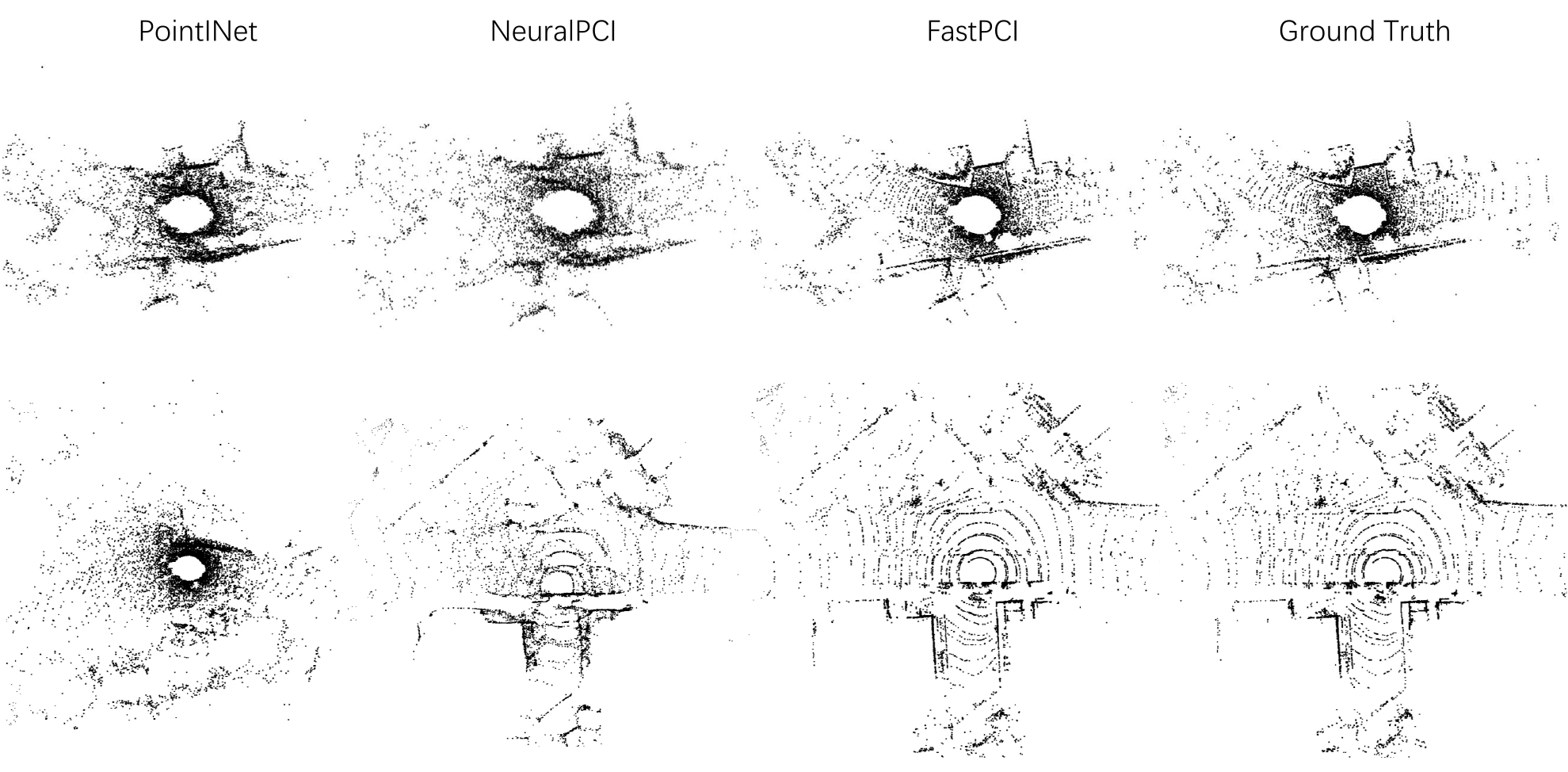}
\includegraphics[page=2,width=1.0\columnwidth]{figures/src/figure_S4.pdf}
\caption{\textbf{More qualitative results.}
}
\label{fig:figureS4}
\end{figure}

\begin{figure}[t]
\centering
\includegraphics[trim={0, 0, 0, 0}, clip, width=1.0\columnwidth]{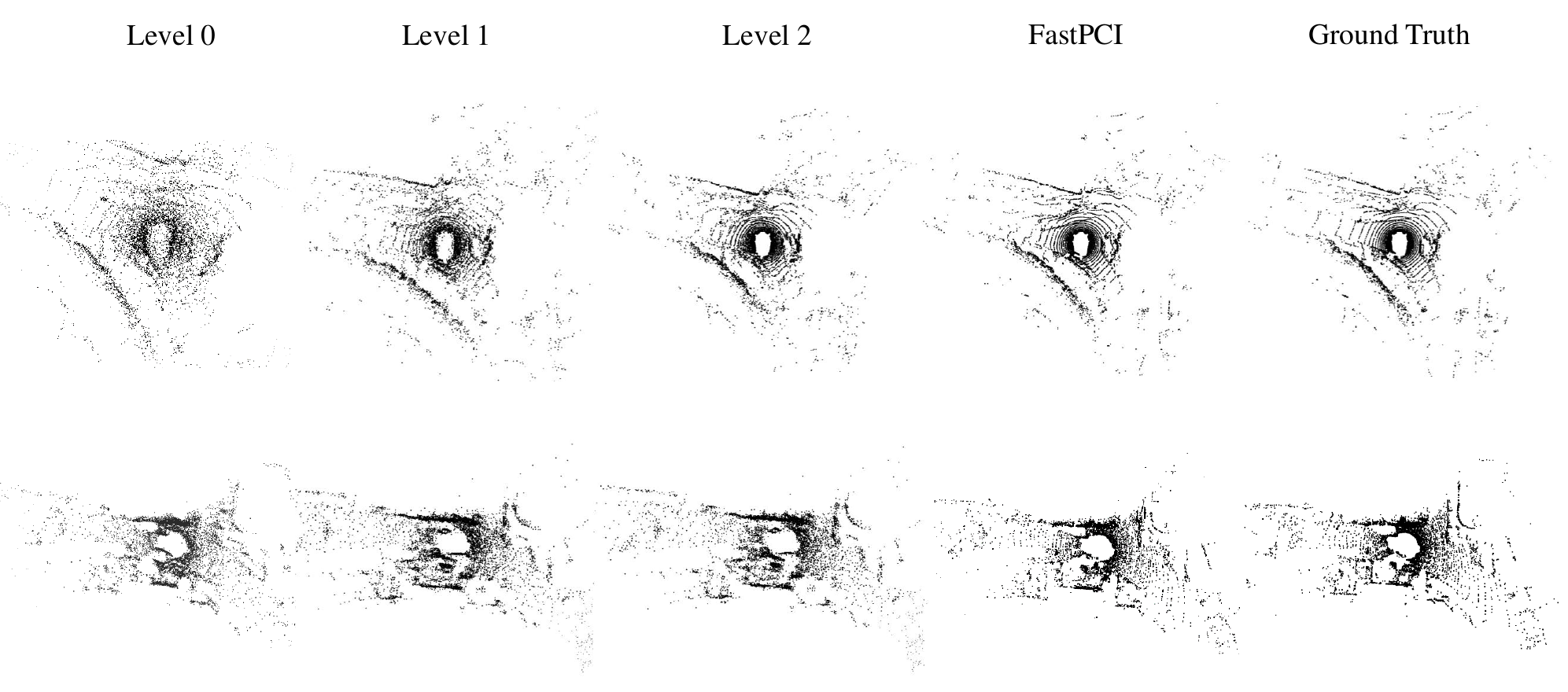}
\caption{\textbf{Visualization of Pyramid results of our FastPCI.} 
}
\label{fig:figureS5}
\end{figure}
\begin{figure}[t]
\centering
\includegraphics[trim={0, 0, 0, 0}, clip, width=0.8\columnwidth]{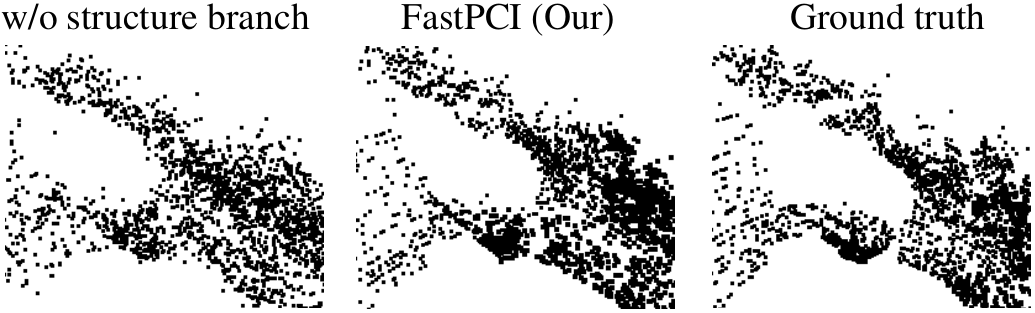}
\caption{\textbf{Qualitative ablation study of structure branch.}
}
\label{fig:rebut1}
\end{figure}

\section{Methods Details}
\label{sec:supp_method}

\figlabel\ref{fig:figureS1} shows the architecture for the Pyramid Motion-Structure Estimation Network. \figlabel\ref{fig:figureS2} shows the Motion Compensation block. Wherein, cost volume represents the cost of scene flow per point. Concatenating with scene flow is reasonable and empirically works. \figlabel\ref{fig:figureS3} shows the details of RefineNet. Our model is small and consists of only 5M parameters.

For the coordinate map $B$, three evenly spaced one-dimensional vectors ranging from 0 to 1 for each pyramid level are generated respectively. The size of each vector equals the number of points in the point cloud of this level. These one-dimensional vectors are then cascaded into a three-dimensional vector $B$, corresponding to positions $x$, $y$, and $z$ respectively.

\section{Additional Ablation Study}\label{sec:supp_exp}


\noindent\textbf{Investigation of Dual Direction Cross Attention and Self Attention.} To verify our proposed Dual-Direction Cross-Attention in the point cloud frame interpolation task, we conduct a comparison experiment with the self-attention mechanism, as shown in Table \ref{table:ablate_dualdir}. The self-attention mechanism performs self-attention in each direction without cross-attention between directions. The results show that our Dual-Direction Cross-Attention improves the performance by nontrivial margins.


\begin{table*}[t]
    \centering
    \caption{\textbf{Investigation of Dual-Direction Cross-Attention and Self-Attention} on all three datasets. The Dual-Direction Cross-Attention module is modified by excluding the input of the reverse feature to compare it with the Self-Attention mechanism. This modification enables us to establish a self-attention experiment and evaluate its performance.
    }
    \label{table:ablate_dualdir}
    {\begin{tabular}{l|c|cc}
    \toprule
    \multirow{2}{*}{\textbf{Dataset}} & \multirow{2}{*}{\textbf{Component}} & \multicolumn{2}{c}{\textbf{Metrics}} \\ 
    && \textbf{CD$\downarrow$} & \textbf{EMD$\downarrow$} \\ \midrule
    \multirow{2}{*}{KITTI odometry}&{Self-Attention}&{0.59}&{55.48} \\
    &{\textbf{Dual-Direction Cross-Attention (Ours)}}& \textbf{0.58}&\textbf{53.78} \\\midrule
    \multirow{2}{*}{Argoverse 2 sensor}&{Self-Attention}&{0.78}&{60.55} \\
    &{\textbf{Dual-Direction Cross-Attention (Ours)}}& \textbf{0.73}&\textbf{57.95} \\ \midrule
    \multirow{2}{*}{Nuscenes}&{Self-Attention}&{1.43}&{185.86} \\
    &{\textbf{Dual-Direction Cross-Attention (Ours)}}& \textbf{1.11} & \textbf{173.64} \\ \midrule
    \end{tabular}
    }
    \vspace{-1.0em}
\end{table*}


\section{Additional Visual Results}\label{sec:supp_visual}

\noindent\textbf{More Qualitative Results.}
\figlabel\ref{fig:figureS4} shows more of our qualitative results.

\noindent\textbf{Pyramid Results.}
\figlabel\ref{fig:figureS5} shows our results in different levels from the pyramid structure.

\noindent\textbf{Evidence on structure preservation of Motion-Structure block.}
Our structure awareness is optimized end-to-end by the point cloud reconstruction through the final layer and from the different hierarchies in our pyramid network (\textit{\underline{\S 3.2}}). 
The effectiveness of the structure branch is ablated in \textit{\underline{Table 3}}. We provide a qualitative ablation study in Figure \ref{fig:rebut1}, which clearly shows the structure preservation with our motion-structure block.

\clearpage



%
%
\bibliographystyle{splncs04}
\bibliography{egbib}